\documentclass[runningheads]{llncs}

\usepackage{eccv}

\usepackage{eccvabbrv}
\usepackage{wrapfig}

\usepackage{graphicx}
\usepackage{multirow}
\usepackage{booktabs}

\usepackage[accsupp]{axessibility}  

\usepackage{hyperref}

\usepackage{orcidlink}

\definecolor{gray}{RGB}{146,146,146}
\definecolor{light}{RGB}{204,206,10}
\definecolor{dark}{RGB}{60,120,141}

\begin{document}

\title{Pseudo-Labeling by Multi-Policy Viewfinder Network for Image Cropping} 

\titlerunning{Pseudo-Labeling by Multi-Policy Viewfinder Network for Image Cropping}

\author{Zhiyu Pan \and
Kewei Wang \and
Yizheng Wu \and
Liwen Xiao \and
Jiahao Cui \and
Zhicheng Wang \and
Zhiguo Cao
}

\authorrunning{Z.~Pan et al.}

\institute{Key Laboratory of Image Processing and Intelligent Control, Ministry of Education School of Artificial Intelligence and Automation, Huazhong University of Science and Technology, Wuhan, 430074, China. \\
\email{\{zhiyupan;zgcao\}@hust.edu.cn}}

\maketitle

\begin{abstract}
  Automatic image cropping models predict reframing boxes to enhance image aesthetics. Yet, the scarcity of labeled data hinders the progress of this task. To overcome this limitation, we explore the possibility of utilizing both labeled and unlabeled data together to expand the scale of training data for image cropping models. This idea can be implemented in a pseudo-labeling way: producing pseudo labels for unlabeled data by a teacher model and training a student model with these pseudo labels. However, the student may learn from teacher's mistakes. To address this issue, we propose the multi-policy viewfinder network (MPV-Net) that offers diverse refining policies to rectify the mistakes in original pseudo labels from the teacher. The most reliable policy is selected to generate trusted pseudo labels. The reliability of policies is evaluated via the robustness against box jittering. The efficacy of our method can be evaluated by the improvement compared to the supervised baseline which only uses labeled data. Notably, our MPV-Net outperforms off-the-shelf pseudo-labeling methods, yielding the most substantial improvement over the supervised baseline. Furthermore, our approach achieves state-of-the-art results on both the FCDB and FLMS datasets, signifying the superiority of our approach.
\end{abstract}

\section{Introduction}
\label{sec:intro}

\begin{figure}
  \centering
  \subcaptionbox{Vanilla pseudo-labeling for the image cropping task and the corresponding evolution trajectory of pseudo labels.}{\includegraphics[width=0.48\columnwidth]{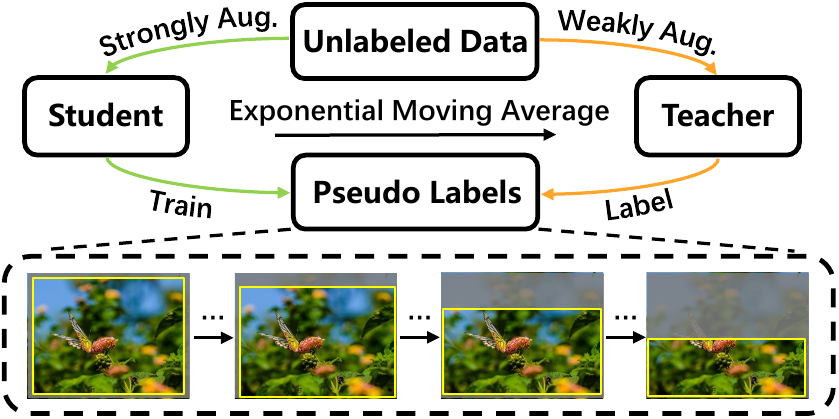}}
  \subcaptionbox{Rectifying pseudo labels with the proposed MPV-Net and the corresponding evolution trajectory of the rectified pseudo labels.}{\includegraphics[width=0.48\columnwidth]{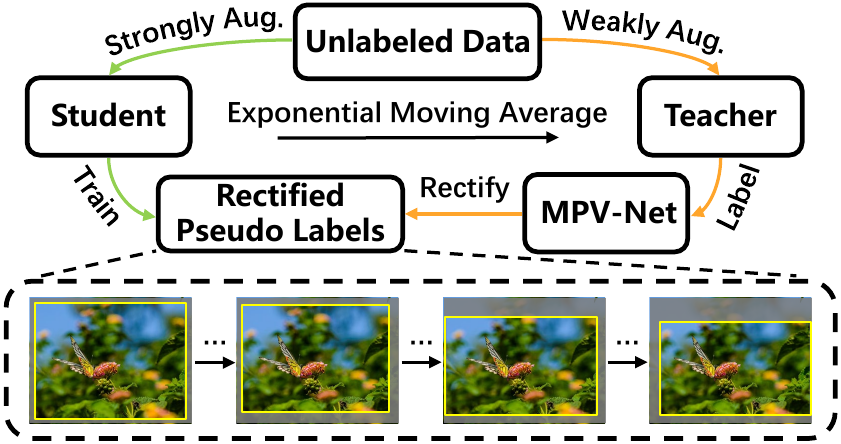}}
  \vspace{-6pt}
  \caption{\textbf{The role that MPV-Net plays in pseudo-labeling for image cropping.} (a) With vanilla pseudo-labeling, the student may learn from mistakes of the teacher, which is called the problem of confirmation bias~\cite{tarvainen2017mean}. The confirmation bias may cause a deterioration of pseudo labels during the iteration. (b) We propose the multi-policy viewfinder network (MPV-Net) to rectify the mistakes of teacher, which makes the pseudo label trusted.}
  \vspace{-16pt}
  \label{fig:fig1}
\end{figure}

Automatic image cropping~\cite{islam2017survey} aims to enhance aesthetic quality of images by cropping with predicted reframing boxes. Reframing box annotations requires expertise in aesthetics~\cite{chen2017quantitative}, which limits annotators to the professional population. This characteristic makes it challenging to collect a large dataset with sufficient labels. Recent regression-based image cropping methods~\cite{hong2021composing,jia2022rethinking,lu2020learning} make efforts to address the scarcity of labeled data with extra data in different annotation forms from other related tasks~\cite{lee2018photographic,zeng2019reliable,zeng2020grid}. However, the amount and domain gap of extra data are still limiting factors. 
Our work tries to address this problem by leveraging unlabeled data to facilitate learning with all available labeled data.


Learning with all available labeled data and extra unlabeled data falls within the realm of omni-supervised learning~\cite{radosavovic2018data}, a special type of semi-supervised learning~\cite{van2020survey,zhu2009introduction}.
The common solution of recent omni-/semi- supervised methods is pseudo-labeling~\cite{sohn2020fixmatch,tarvainen2017mean}, which has shown effectiveness in various computer vision tasks, \eg, object detection~\cite{chen2022label,liu2021unbiased} and semantic segmentation~\cite{fan2022ucc,qiao2023fuzzy}. In the vanilla pseudo-labeling framework, as illustrated in Fig.~\ref{fig:fig1}(a), a student model learns from the pseudo labels on strongly augmented unlabeled data; and a teacher model, exponential moving average of the student, generates pseudo labels for corresponding weakly augmented unlabeled data. This training strategy may suffer from the confirmation bias~\cite{tarvainen2017mean}: the student model learns from the mistakes of the teacher model, which hinders the learning of necessary knowledge. The problem of confirmation bias results in a deterioration of pseudo labels. 



The confirmation bias can be mitigated by improving the quality of pseudo labels~\cite{tarvainen2017mean}.
Previous methods~\cite{jeong2019consistency,sohn2020simple} guarantee a high quality of pseudo labels by selecting reliable pseudo labels according to confidence scores. However, the regression-based image cropping models predict a reframing box without any confidence indicator~\cite{hong2021composing,pan2023find}, which makes it an open problem to search for better pseudo labels in our case. 
So, we argue to hunt for better pseudo labels by learning,
as illustrated in Fig.~\ref{fig:fig1}(b), we propose to train an auxiliary network to rectify the mistakes in original pseudo labels.

Inspired by that expert photographers search for the optimal perspective by rectifying their framing polices through the viewfinder repeatedly, we propose a multi-policy viewfinder network (MPV-Net) which provides diverse rectifying policies via independent regression heads to correct pseudo labels generated from the teacher model. 
The most reliable rectifying policy is selected to generate trusted pseudo labels.  
We hypothesize that a reliable rectifying policy is committed to its own results. Hence, a stability-based policy selecting method is proposed: the input of MPV-Net is jittered several times, then each policy will generate a bunch of rectified results. The variance~\cite{xu2021end} of rectified results from each policy is used to measure its reliability. The rectified pseudo label from the policy with the lowest variance is selected as the trusted one. With the MPV-Net and policy selecting mechanism, as shown in Fig.~\ref{fig:fig1}(b), the rectified pseudo labels evolve to be trusted. More comparison of pseudo label evolution trajectories is in Fig.~\ref{fig:ablation}.

The efficacy of our method can be evaluated by the improvement compared to the supervised baseline that only uses the labeled data~\cite{jeong2019consistency,radosavovic2018data,sohn2020fixmatch,sohn2020simple,tarvainen2017mean,zhou2021instant}. Our framework achieves the most substantial improvement in contrast with off-the-shelf pseudo-labeling methods.
Besides, we report new state-of-the-art (SOTA) performance over other regression-based image cropping methods on FCDB~\cite{chen2017quantitative} and FLMS~\cite{fang2014automatic} datasets. 
The contributions of our work are as follows:
\begin{itemize}
    \item[$\bullet$] We firstly present the pseudo-labeling framework to mitigate the scarcity of annotations for image cropping;
    \item[$\bullet$] We design the MPV-Net with a policy selecting method to address the confirmation bias of vanilla pseudo-labeling;
    \item[$\bullet$] We achieve new SOTA performance on FCDB and FLMS datasets.
\end{itemize}

\section{Related Work}

\textbf{Automatic image cropping} algorithms rely on manually defined rules in early phases~\cite{chen2003visual,cheng2010learning,marchesotti2009framework,nishiyama2009sensation,santella2006gaze,suh2003automatic,yan2013learning,zhang2005auto}. Since aesthetics is too abstract to be defined as rules, data-driven approaches have become mainstream. The annotation of reframing boxes requires expertise in aesthetics, which makes it expensive to collect a large scale labeled dataset~\cite{chen2017quantitative}. So, prior arts transform the automatic image cropping to the candidate-selection paradigm~\cite{wei2018good} and label different views in one image with mean opinion scores (MOS) through crowdsourcing~\cite{zeng2019reliable,zeng2020grid}. The reframed result of the candidate-selection methods~\cite{chen2017learning,pan2021transview,wang2023image,wang2018deep,wei2018good} is the candidate with the highest MOS estimation. However, the final results of these candidate-selection methods are highly dependent on the predefined candidates, which limits real-world practice~\cite{pan2023find}. Therefore, some works~\cite{li2018a2,li2019fast} predict the reframing box boundaries step by step based on reinforcement learning~\cite{li2017deep}. Due to the ambiguity of the reward function in reinforcement learning, recent works try to directly learn from the reframing box annotations and predict the boxes by regression~\cite{hong2021composing,jia2022rethinking,lu2020learning,pan2023find}. Most of these approaches rely on auxiliary labeled data from related tasks~\cite{greco2013saliency,lee2018photographic} to address the scarcity of reframing box labels. However, the amount and the domain gap of the extra labeled datasets still limit the performance. So, our work exploits unlimited cheap unlabeled data to address this problem from the perspective of pseudo-labeling.

\textbf{Omni-/semi-supervised learning} methods~\cite{berthelot2019mixmatch,grandvalet2004semi,iscen2019label,lee2013pseudo,rasmus2015semi,sajjadi2016regularization} delve into the problem of how to make use of both labeled and unlabeled data. Omni-supervised learning~\cite{radosavovic2018data} where all available labeled data are expected to be used and the unlabeled data are also included to achieve better performance than the supervised learning; semi-supervised learning~\cite{van2020survey} focuses on how to achieve considerable performance with limited annotations. 
Recent omni-/semi-supervised learning methods~\cite{laine2016temporal,tarvainen2017mean} follow a pseudo-labeling paradigm where a student learns from the pseudo labels generated from the teacher. These approaches achieve remarkable improvement for many computer vision tasks, \eg, object detection~\cite{chen2022label,tang2021humble,xu2021end,yang2021interactive} and semantic segmentation~\cite{fan2022ucc,qiao2023fuzzy}. However, when it is applied to new tasks, vanilla pseudo-labeling method, \eg, the mean teacher~\cite{tarvainen2017mean}, may have struggles because of the confirmation bias. The confirmation bias means the student might learn from the mistakes of teacher. Existing methods try to alleviate the influence of confirmation bias by selecting from pseudo labels~\cite{sohn2020simple,liu2022unbiased}. Different from these methods, our work rectify pseudo labels via the proposed MPV-Net, which ensures a higher upper-bound of the pseudo label quality.
Some previous image cropping methods also try to use unlabeled images. However, they either generate pseudo training pairs on curated data~\cite{chen2017learning} or require unlabeled data to be included in test samples~\cite{wang2023image}. Different from these methods, our method is more flexible: there is no restriction for unlabeled data.

\section{Methodology}

In our method, the labeled data $D_l = \{(x^i_l,y^i_l)\}_{i=1}^{n_l}$ and the unlabeled data $D_u = \{x^i_u\}_{i=1}^{n_u}$ are given, where $x$ and $y$ represent the image and the reframing box label respectively, $n_l$ and $n_u$ are the total number of labeled and unlabeled data. With both $D_l$ and $D_u$, it is expected to train a reframing box regression model that can outperform the one only trained on $D_l$, \ie, the supervised baseline. We firstly provide an overview of how to take advantage of $D_u$, then introduce how to address the confirmation bias in image cropping task with the multi-policy viewfinder network (MPV-Net) and the policy selecting method.

\subsection{Overview}
\begin{figure*}[!t]
  \centering
  \includegraphics[width=\textwidth]{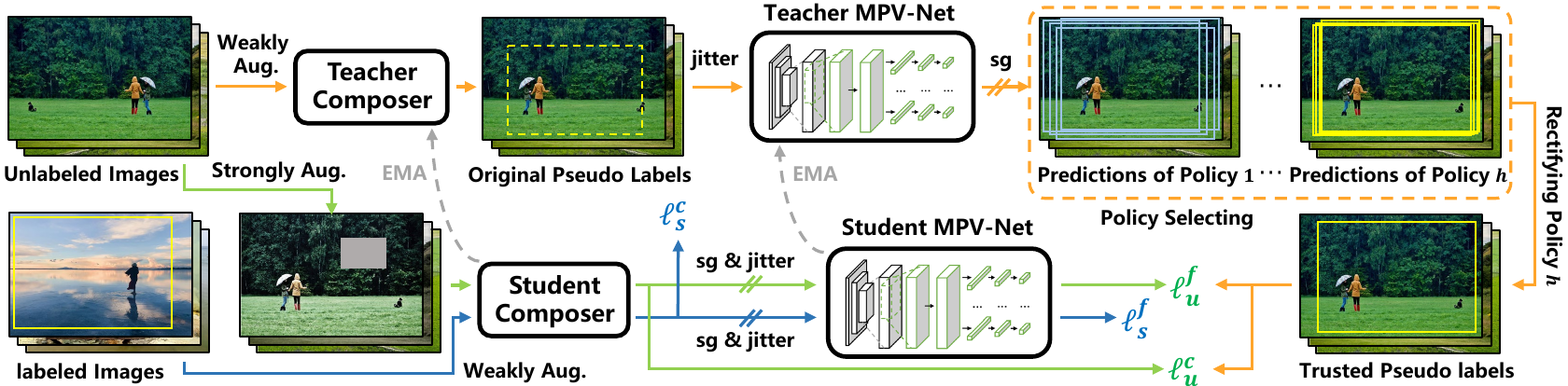}
  \vspace{-16pt}
  \caption{\textbf{The technical pipeline of the proposed pseudo-labeling framework for image cropping.} In this work, the teacher branch includes the reframing box regression model, \textit{a.k.a.} the composer, and the multi-policy viewfinder network (MPV-Net). Each head of the MPV-Net can provide with a rectifying policy for the original pseudo labels predicted by the teacher composer. The proposed policies are selected according to their stability. The most stable policy is used to generate the trusted pseudo labels. The teacher models are updated by their student models respectively via exponential moving average (EMA)~\cite{tarvainen2017mean}. The student models are optimized by both the supervised and unsupervised losses. ``sg'' in the figure means stopping gradient.}
  \label{fig:pipeline}
  \vspace{-8pt}
\end{figure*}

As illustrated in Fig.~\ref{fig:pipeline}, the proposed method follows the teacher-student pseudo-labeling framework. The teacher branch generates pseudo labels on the unlabeled images for the training of the student branch. Some deficient pseudo labels may lead to the confirmation bias. So, we try to rectify the pseudo labels with an extra network. We design the MPV-Net to propose different rectifying policies. Each policy is equipped with unique knowledge via independent training. A policy selecting mechanism is invented to select the most reliable policy that produces the trusted pseudo label. In our framework, the CACNet~\cite{hong2021composing} serves as the default reframing box regression model, which is called the composer in our paper. The teacher $\tau$ (teacher composer $\tau^c$ and teacher MPV-Net $\tau^f$) is updated by the student $\xi$ (student composer $\xi^c$ and student MPV-Net $\xi^f$) via the exponential moving average (EMA)~\cite{tarvainen2017mean}:
\begin{equation}
    \tau_i \leftarrow \alpha\tau_{i-1} + (1-\alpha)\xi_i\,,
\end{equation}
where $\tau_i$ and $\xi_i$ are the parameters of the teacher and student in the $i$-th iteration respectively, and $\alpha$ is the hyper-parameter to control the updating speed. The student is optimized by the combination of supervised loss for labeled images and unsupervised loss for unlabeled images. For student composer $\xi^c$, the supervised loss $\ell_s^c$ and unsupervised loss $\ell_u^c$ can be calculated as:  
\begin{gather}
    \ell_s^c = \frac{1}{n_l}\sum_{i=1}^{n_l}\ell_1(\xi^c(x^i_l), y^i_l)\,,\\
    \ell_u^c = \frac{1}{n_u}\sum_{i=1}^{n_u}\ell_1(\xi^c(x^i_u), \hat{y}^i_u)\,,
\end{gather}
where $\xi^c$ is the parameter of student composer, $\hat{y}^i_u$ is the pseudo label for the $i$-th unlabeled image $x^i_u$, and $\ell_1(\cdot)$ is $\ell_1$ loss. 
The supervised loss $\ell_s^f$ and the unsupervised loss $\ell_u^f$ for the student MPV-Net (defined in Section~\ref{subsec:rectifier}) are integrated with $\ell_s^c$ and $\ell_u^c$ for the student composer, hence the overall loss can be computed as:
\begin{equation}
    \ell = \ell_s^c + \ell_s^f + \lambda(\ell_u^c + \ell_u^f)
\end{equation}
where $\lambda$ is a hyper-parameter used to trade off between the supervised loss and the unsupervised loss. 

\subsection{Multi-Policy Viewfinder Network}
\label{subsec:rectifier}

\begin{figure}[!t]
  \centering
  \includegraphics[width=0.67\textwidth]{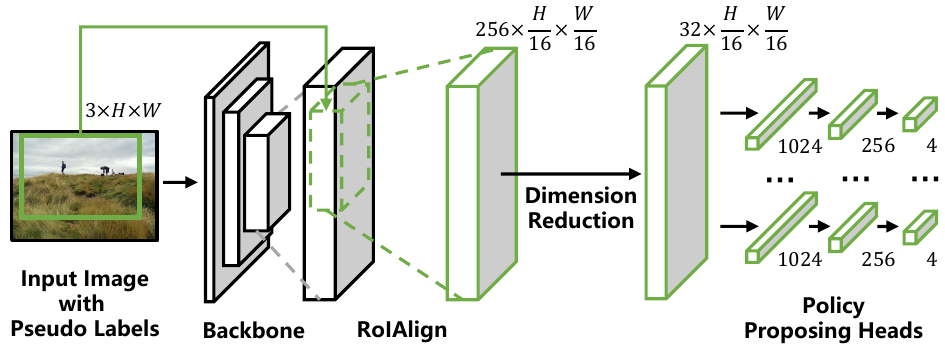}
  \vspace{-8pt}
  \caption{\textbf{The architecture of the MPV-Network.} The green box is the original pseudo label. The multiple heads provide different rectifying policies for the original pseudo labels. The heads are initialized and trained independently.}
  \label{fig:proposer}
  \vspace{-8pt}
\end{figure}
Even for expert photographers, it is hard to get a good shot all at once, which motivates us to design an extra network to rectify the pseudo labels for the goal of addressing the problem of confirmation bias.
Observing that expert photographers adjust their framing by multiple attempts, our MPV-Net is designed to have multiple heads generating diverse rectifying policies.

As shown in Fig.~\ref{fig:proposer}, the inputs of MPV-Net are images with the corresponding pseudo labels generated from the composer; the outputs of MPV-Net are four offsets for left, upper, right, and bottom boundaries of the original pseudo labels. MPV-Net is composed of a feature encoder, a RoIAlign process~\cite{he2017mask} used to extract the feature in the pseudo reframing box, and $h$ independently initialized policy proposing heads. For training the student MPV-Net, the detached predicted reframing boxes from the student composer are randomly jittered $h$ times to provide varied training samples for $h$ heads, respectively. In this way, different heads learn from independent training samples and are endowed with different knowledge for different rectifying policies. To make sure the MPV-Net can appropriately fit the distribution of the pseudo labels from the teacher branch, the range of the box-jittering is gradually reduced at the training stage. The supervised loss $\ell_s^f$ and the unsupervised loss $\ell_u^f$ of the student MPV-Net $\xi^f$ are:
\begin{gather}
    \ell_s^f = \frac{1}{n_l}\sum_{i=1}^{n_l}\frac{1}{h}\sum_{j=1}^h\ell_1(\xi^f(x^i_l,b^i_j|\phi_j)+b^i_j, y^i_l)\,,\\
    \ell_u^f = \frac{1}{n_u}\sum_{i=1}^{n_u}\frac{1}{h}\sum_{j=1}^h\ell_1(\xi^f(x^i_u,b^i_j|\phi_j)+b^i_j, \hat{y}^i_u)\,,
\end{gather}
where $\xi^f(\cdot|\phi_j)$ is the student MPV-Net conditioned on the $j$-th proposing head $\phi_j$, $b^i_j=\delta(\xi^c(x^i))$ is the noisy pseudo label to be rectified for the $j$-th head, and $\delta(\cdot)$ is the random box-jittering process. When the MPV-Net serves as a part of the teacher, the most reliable policy for the unlabeled image is selected by the policy selecting method to generate the most trusted pseudo label.

\subsection{Policy Selecting}
\label{sec:ps}

\begin{figure}
  \centering
  \includegraphics[width=0.95\columnwidth]{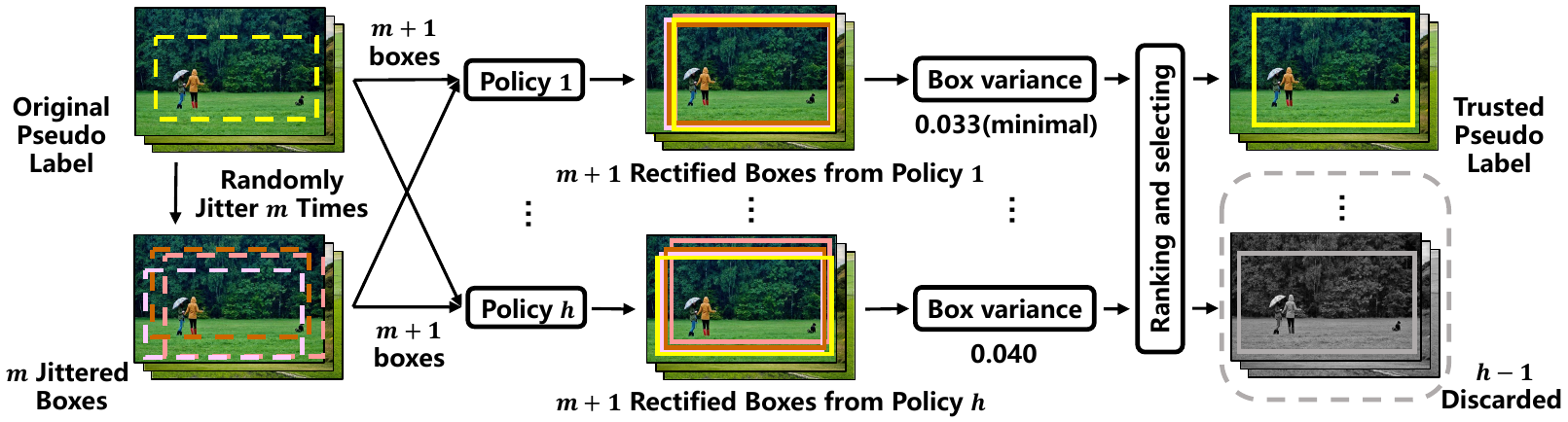}
  \vspace{-5pt}
  \caption{\textbf{The policy selecting mechanism.} The original pseudo labels from the teacher composer are jittered $m$ times. The jittered boxes and the original box are fed into the teacher MPV-Net. Each policy of the MPV-Net generates a set of rectified boxes. Then we calculate the variance~\cite{xu2021end} of the rectified boxes from each policy to measure its stability. The policies are ranked based on their variance. We select the policy with the lowest variance to rectify the original pseudo label as the trusted one. The solid boxes are the rectified results of the dashed boxes in the same color.}
  \label{fig:selection}
  \vspace{-8pt}
\end{figure}

Prior image cropping solutions~\cite{zeng2019reliable,zeng2020grid} observe the local smoothness of reframed results. This is also consistent with that a good cropping model can tolerate local perturbation~\cite{pan2021robust}. According to these observations of previous image cropping methods, we hypothesize that the disturbance-stable rectifying policies are more reliable to generate trusted pseudo labels. This hypothesis is statistically validated by a verification experiment in Section.~\ref{sec:analysis}. 

To measure the stability of heads, \textit{a.k.a.} the rectifying policies, we evaluate the consistency of their rectified results when the original pseudo labels are added noise. In practice, for each unlabeled image, we jitter the original pseudo labels from the teacher composer for $m$ times. Specifically, as shown in Fig.~\ref{fig:selection}, we can obtain the original pseudo label $p^i$ for the $i$-th unlabeled image $x^i_u$ by: 
\begin{equation}
    p^i = \tau^c(x^i_u)\,,
\end{equation}
where $\tau^c$ is the teacher composer. After jittering $p^i$ for $m$ times and feeding $p^i$ with $m$ jittered boxes into the teacher MPV-Net $\tau_f$, then the $j$-th head $\phi_j$ of the teacher MPV-Net generates a set of rectified results $R_j=\{r_z|z=0,1,...,m\}$, in which
\begin{gather}
    r_0 = \tau^f(x^i_u,p^i|\phi_j)+p^i\,,\\
    \underset{z=1,...,m}{r_z} = \tau^f(x^i_u,\delta_z(p^i)|\phi_j)+\delta_z(p^i)\,,
\end{gather}
where $\delta_z(\cdot)$ is the $z$-th box-jittering process.
Then we can calculate the box regression variance $\sigma_j$ of the rectified results $R_j$ from the $j$-th head as:
\begin{equation}
    \sigma_j = \frac{1}{4}\sum_{k=1}^4 \frac{\overline{\sigma}_k}{0.5(\eta(r_0)+\pi(r_0))} \,,
\end{equation}
where $\overline{\sigma}_k$ indicates the standard deviation of the $k$-th coordinate of the boxes in $R_j$, $\eta(\cdot)$ and $\pi(\cdot)$ are the height and the width of the boxes in $R_j$. Then we can select the head with lowest box variance as the most reliable one. The trusted pseudo label $\hat{y}^i_u$ for the $i$-th unlabeled image $x_i^u$ is:
\begin{gather}
    \hat{y}^i_u = r_0, r_0 \in R_v \,,\\
    v = \arg\min(\{\sigma_j|j=1,...,h\})\,,
\end{gather}
where $\arg\min(\cdot)$ returns the index of the smallest element in the set, and $h$ is the total number of the heads in the MPV-Net. The pseudo labels are updated once per epoch. Different from \cite{xu2021end}, our work uses the box variance to evaluate the rectifying policies rather than the candidates.

\subsection{Implementation Details}

The data augmentations include the composition-irrelevant operations: horizontal flipping, histogram equalization, color inversion, smoothing, posterization, image enhancement, and cutout~\cite{devries2017improved}. Strong data augmentation includes all these operations. Weak data augmentation contains only the horizontal flipping. Validation experiments and visualizations can be found in supplementary materials.

The composer is the CACNet~\cite{hong2021composing} without multi-task training. The CACNet re-weights the predictions with the activation map from the multi-task branch. In our setting, labeled data of other tasks should not be included. Hence, we replace the activation map with a proposal branch which dynamically selects the informative anchors. Details can be found in supplementary materials.

The training process of our omni-supervised framework contains two stages: the warm-up stage and the semi-supervised stage. At the warm-up stage, the student only learns from the labeled data. When it switches to the semi-supervised stage, the teacher copies the weights of the student as the initial weights.

The inference process uses the teacher composer and the most reliable head of the teacher MPV-Net to generate the reframing boxes for the test samples. The teacher composer alone can also achieve similar performance, which is illustrated in Section.~\ref{sec:analysis}. This shows that the performance improvement does not come from larger model parameters.

\section{Experiments}
\subsection{Datasets and Metrics}
We evaluate our framework on the FCDB~\cite{chen2017quantitative} and FLMS~\cite{fang2014automatic} datasets. The FCDB dataset contains $1,743$ images. Each of them has one reframing box annotation. $1,395$ images of the FCDB dataset are used for training, and $348$ images are for testing. For analysis, we follow the setting of CACNet~\cite{hong2021composing} to build the validation set which contains $200$ images. The FLMS dataset contains $500$ images. Each of them has $10$ reframing box annotations from different annotators. All the images of the FLMS dataset are used for testing. We collect the unlabeled data from the CPC~\cite{wei2018good}, CUHK-ICD~\cite{yan2013learning}, and GAIC~\cite{zeng2019reliable} datasets. These datasets provide MOS annotations but no reframing box labels. The CPC dataset contains $12,568$ images; the CUHK-ICD dataset has $1,113$ images; the GAIC dataset owns $3,336$ images. Combining all these images, we obtain $17,017$ unlabeled images for training. Following the setting of previous works, we report the average intersection over union (IoU) and boundary displacement error (BDE) as performance evaluation metrics on the FCDB test set. As for the FLMS dataset, the average maximum IoU and minimum BDE over multiple annotations are used for evaluation. 

\subsection{Experimental Settings}
The input images are resized to $224\times224$. The framework is trained for $60$ epochs, in which the first $9$ epochs are for the warm-up training. The backbone is VGG16~\cite{simonyan2014very}. The model size of composer is $15.7$M. The optimizer for the student is the Adam~\cite{kingma2014adam}. The learning rate for the first $30$ epochs is $3e-4$, and it decays to the half in the remaining epochs. The number of heads is set as $h=5$. We use $\alpha=0.995$ for the EMA, and $\lambda=4$ for the unsupervised loss. The batch size is set to $64$. Each batch includes randomly sampled labeled and unlabeled images. The training takes around $16$ hours.

\subsection{Performance Comparison}
\label{comparison}

\begin{table}[!t]
\begin{minipage}[!t]{0.47\textwidth} \small
    \footnotesize
    \centering
    \captionof{table}{Quantitative comparison with the omni-supervised baselines. Best performance is in \textbf{boldface}, and the second best is \underline{underlined}.}
    \vspace{-5pt}
    \resizebox{\linewidth}{!}{
    \begin{tabular}{lcccc}
    \toprule
    \multirow{2}*{Approaches} &\multicolumn{2}{c}{FCDB} &\multicolumn{2}{c}{FLMS}\\
    & IoU$\uparrow$ &BDE$\downarrow$ &IoU$\uparrow$ &BDE$\downarrow$\\
    \hline
    Supervised &0.699 &0.074 &0.831 &0.039\\
    \hline
    CSD~\cite{jeong2019consistency} &0.700 &0.075 &0.833 &0.039\\
    STAC~\cite{sohn2020simple} &\underline{0.713} &\underline{0.071} &0.843 &\underline{0.035}\\
    InstantTeaching~\cite{zhou2021instant} &0.703 &0.074 &0.833 &0.038\\
    SoftTeacher~\cite{xu2021end} &0.709 &0.073 &0.842 &0.036\\
    UnbiasedV$2$~\cite{liu2022unbiased} &0.704 &0.072 &0.841 &0.037\\
    CrossRectify~\cite{ma2023crossrectify} &0.710 &~\underline{0.071} &~\underline{0.846} &~\underline{0.035}\\
    \hline
    Ours &\textbf{0.724} &\textbf{0.068} &\textbf{0.854} &\textbf{0.032}\\ 
    \bottomrule
    \end{tabular}}
    \label{tab:com2semi}
\end{minipage}
\begin{minipage}[!t]{0.5\textwidth} \small
    \centering
    \captionof{table}{Quantitative comparison with SOTA image cropping methods. Best performance is in \textbf{boldface}, and second best is \underline{underlined}.}
    \vspace{-5pt}
    \resizebox{\linewidth}{!}{
    \begin{tabular}{lccccc}
    \toprule
    \multirow{2}*{Approaches} & Extra &\multicolumn{2}{c}{FCDB} &\multicolumn{2}{c}{FLMS}\\
    & Annotation & IoU$\uparrow$ &BDE$\downarrow$ &IoU$\uparrow$ &BDE$\downarrow$\\
    \hline
    VFN~\cite{chen2017learning} &$\times$ &0.684 &0.084 &- &-\\
    A2RL~\cite{li2018a2} &$\times$ &0.664 &0.089 &0.821 &0.045\\
    A3RL~\cite{li2019fast} &$\times$ &0.696 &0.077 &0.839 &-\\
    AARN~\cite{lu2020learning} &\checkmark &0.673 &- &0.846 &-\\
    CBLNet~\cite{pan2023find} &$\times$ &\underline{0.718} &\underline{0.069} &0.838 &0.040\\
    CAC~\cite{hong2021composing} &$\times$ &0.700 &0.075 &0.839 &0.039\\
    CAC~\cite{hong2021composing} &\checkmark &0.716 &0.069 &\underline{0.853} &\underline{0.033}\\
    SFRC~\cite{wang2023image} &$\times$ &0.695 &0.075 &- &-\\
    \hline
    Ours &$\times$ &\textbf{0.724} &\textbf{0.068} &\textbf{0.854} &\textbf{0.032}\\
    \bottomrule
    \end{tabular}}
    \label{tab:com2crop}
\end{minipage}
\end{table}

\textbf{Comparison against pseudo-labeling baselines.} Since we are the first to perform pseudo-labeling for image cropping, we implement the following pseudo-labeling baselines which can be directly applied to our task and compare with them. All these baselines are in the omni-supervised setting, \ie, all available labeled data are used:

$\bullet$ CSD~\cite{jeong2019consistency} takes the advantage of unlabeled data with consistency regularization. At the training stage, the input images are flipped and fed into the composer. The consistency loss is calculated between the outputs from the original inputs and the flipped ones.

$\bullet$ STAC~\cite{sohn2020simple} is a self-training method. The teacher in STAC is pre-trained on the labeled data and then is used to generated pseudo labels for unlabeled data. The pseudo labels are never updated in the following iterations.

$\bullet$ Instant-teaching~\cite{zhou2021instant} and CrossRectify~\cite{ma2023crossrectify} improve the quality of the pseudo labels by co-training. Two models with the same architecture but different weights help each other to rectify the deficient labels. 

$\bullet$ Soft Teacher~\cite{xu2021end} is transferred to our task by randomly generating candidates around the references and evaluating them with the box variance.

$\bullet$ Unbiased Teacher V$2$~\cite{liu2022unbiased} relocates the boundaries of pseudo labels by uncertainty estimation~\cite{he2019bounding}.

The results are illustrated in Table~\ref{tab:com2semi}. The improvement compared with the supervised baseline, which is the composer only trained with all labeled data, can represent the efficacy of using unlabeled data. One can observe that training with unlabeled data helps reframing box regression. Compared with off-the-shelf pseudo-labeling learning baselines, our method shows the obvious superiority. The advantage of our method may come from that the proposed MPV-Net obeys the philosophy of photographing.

\textbf{Comparison with SOTA image cropping methods.} Quantitative results on the FCDB and FLMS datasets are illustrated in Table~\ref{tab:com2crop}. From the results, we can make the following observations:

\textit{(a) Regression is more efficient than reinforcement learning for the reframing box prediction.} Compared with the reinforcement learning methods, \ie, A3RL~\cite{li2019fast}, other regression-based methods show an obvious superiority on both of the FCDB and FLMS datasets. 

\textit{(b) Training with unlabeled data may be more helpful than extra annotations from other tasks.} 
AARN~\cite{lu2020learning} and CAC~\cite{hong2021composing} require the extra labeled data of saliency and composition classification. The scale of extra annotations may even larger than which FCDB training set has. Our framework uses only the unlabeled data and outperforms current regression-based image cropping methods, which demonstrates the effectiveness of omni-supervised learning.

\begin{figure}[!t]
  \centering
  \includegraphics[width=\textwidth]{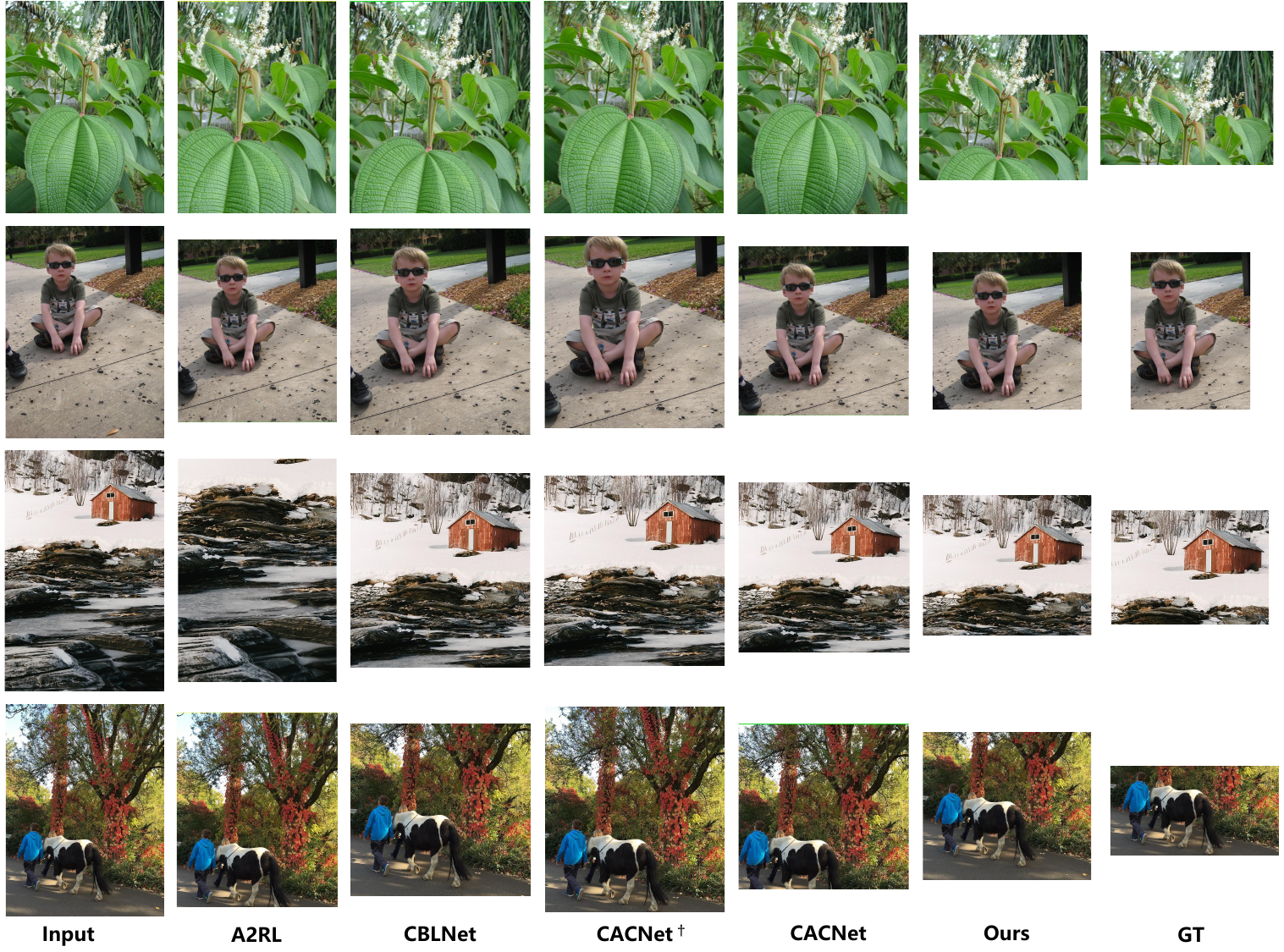}
  \vspace{-18pt}
  \caption{\textbf{Qualitative comparison with SOTA image cropping methods.} Results of CACNet tagged with $\dag$ denotes that the extra annotations from composition classification dataset~\cite{lee2018photographic} are abandoned.}
  \label{fig:qualitativecom}
  \vspace{-8pt}
\end{figure}

Qualitative comparison with other SOTA image cropping methods on FCDB dataset is shown in Fig.~\ref{fig:qualitativecom}. One can draw the following observations:

\textit{(a) Our method can place the foregrounds at proper locations.} If the foregrounds were placed at the edge area, attention of the viewers may be unpleasantly disturbed. When other methods frequently place the foregrounds at the edge area, our method can place the foregrounds at proper locations. Moreover, our method can follow the rule of thirds (the third and forth row). This indicates that our model has the ability to be aware of the concept of beauty.

\textit{(b) Our method keeps less redundancy than other methods.} Too much background or some irrelevant content is the redundancy. The redundancy may distract the viewers and make the foregrounds not prominent. Compared with other methods, our method can keep less redundancy. However, our method tend to keep more background than the ground truths.

\textbf{Subjective evaluation.} Due to the subjective/aesthetic nature of the image cropping task, we conduct a subjective evaluation by the means of user study on FCDB dataset. The questionnaires of our user study are posted on an online survey platform. Each questionnaire presents the reframed results of the supervised baseline, the CACNet, and our method on $30$ randomly sampled images from FCDB dataset and asks the users to choose at least one as the good re-framing solution. $200$ questionnaires are collected, and all the images from FCDB dataset are covered. We use the mean recall rate (how many times are chosen among all trials) as the metrics of the acceptance degree. The results are shown in Table~\ref{tab:userstudy}, one can observe: for the real users, the improvement from leveraging unlabeled data is more obvious. Including unlabeled data in training may be more suitable for real-world practice. Details can be found in supplementary materials.

\begin{table}[!t]
    \footnotesize
    \centering
    \caption{The results of user study on FCDB dataset.}
    \vspace{-6pt}
    \resizebox{0.4\linewidth}{!}{
    \begin{tabular}{ccccc}
    \toprule
    Approaches &Supervised &CACNet &Ours\\
    \hline
    Recall &50.48\% &52.61\% &\textbf{60.38}\%\\
    \bottomrule
    \end{tabular}}
    \label{tab:userstudy}
    \vspace{-8pt}
\end{table}

\subsection{Analyses}
\label{sec:analysis}

\begin{figure}[!t]
  \centering
  \subcaptionbox{The correlation between the IoU and the box regression variance.}{\includegraphics[width=0.49\textwidth]{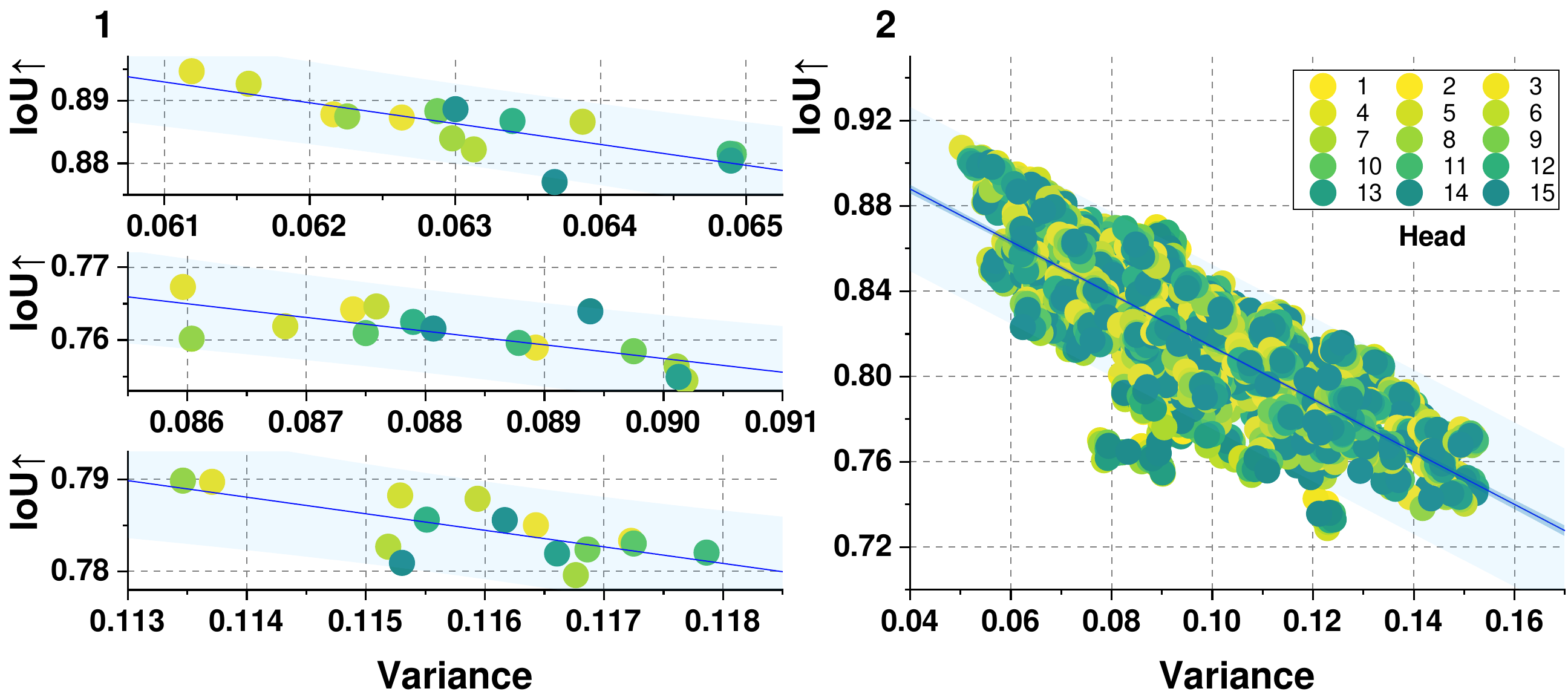}}
  \subcaptionbox{The correlation between the BDE and the box regression variance.}{\includegraphics[width=0.49\textwidth]{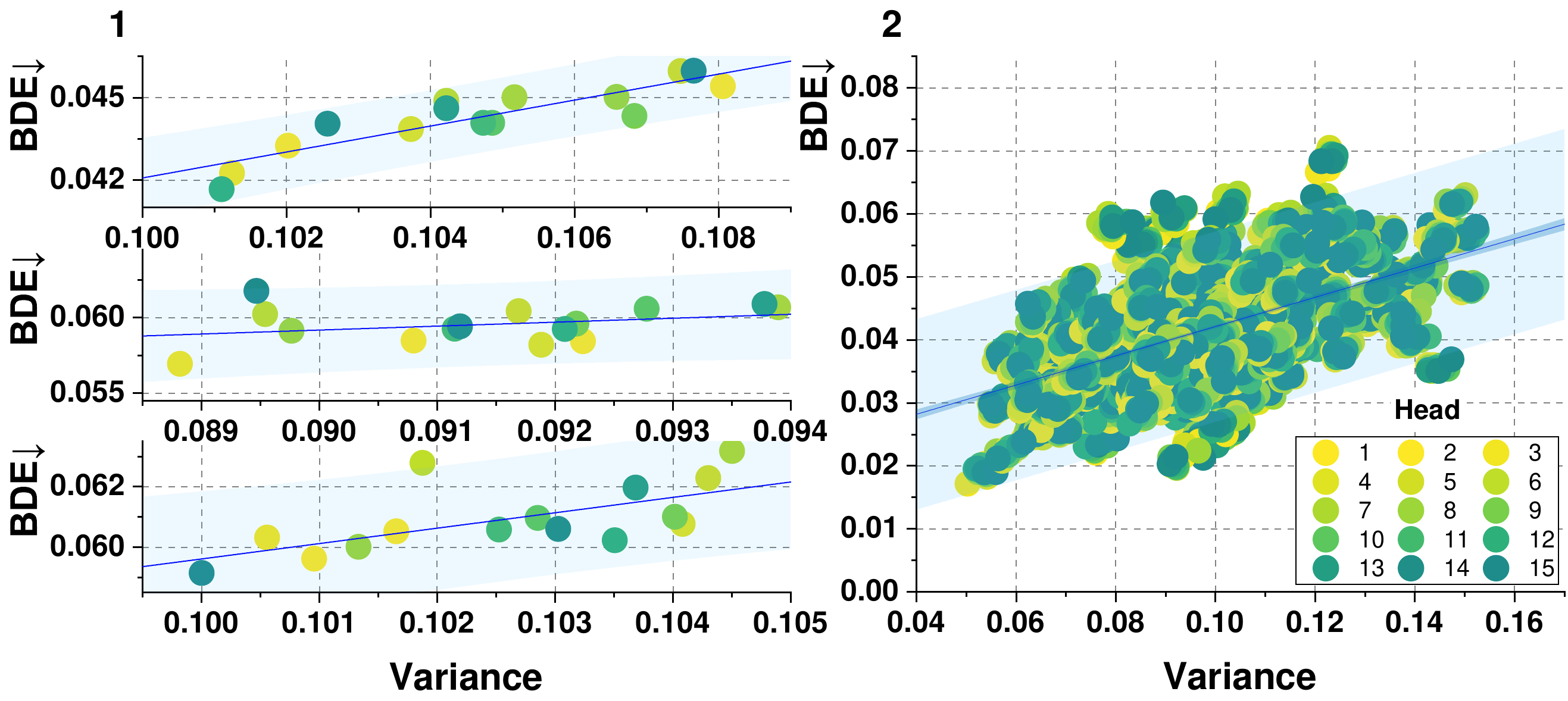}}
  \vspace{-4pt}
  \caption{\textbf{The correlation between the performance and the stability of heads.} The toy MPV-Net with $15$ heads is tested on the FCDB validation set. (a) and (b) show the IoU- and BDE-variance correlation, respectively. The left sub-figures are the results on randomly sampled images; right sub-figures are the overview for the IoU- and BDE-variance correlation on the whole validation set. Points with different colors represent the results from different heads (policies); the blue lines and bars are the linearly fitting results and $95\%$ prediction bands. Best viewed in color.}
  \label{fig:correlation}
  \vspace{-10pt}
\end{figure}

\textbf{Hypothesis verification.} 
To validate the hypothesis: the stable polices can re-frame better, we provide some examples and statistical overviews on FCDB validation set for the correlation of performance and the stability of policy. At the test time, the randomly generated boxes are jittered several times and fed into the toy MPV-Net with $15$ policy proposing heads. Therefore, each policy of the toy MPV-Net can generate multiple re-framing results for the jittered input references. Hence, the box regression variance can be computed for each policy following the process introduced in Sec.~\ref{sec:ps}. The correlation between the variance and the performance of the policy is illustrated in Fig.~\ref{fig:correlation}. As shown in Fig.~\ref{fig:correlation}(a), for the same group of jittered boxes, the policies with the lowest variance in their group achieve the best IoU performance. And on the whole validation set, it seems that there exists a negative correlation between the policy variance and the IoU performance. Results in Fig.~\ref{fig:correlation}(b) show that the BDE performance has a positive correlation with the policy variance. These results validate the design of our policy selecting mechanism.

\begin{figure}[!t]
  \centering
  \subcaptionbox{Influence of the unlabeled data quantity on FCDB dataset.}{\includegraphics[width=0.49\textwidth]{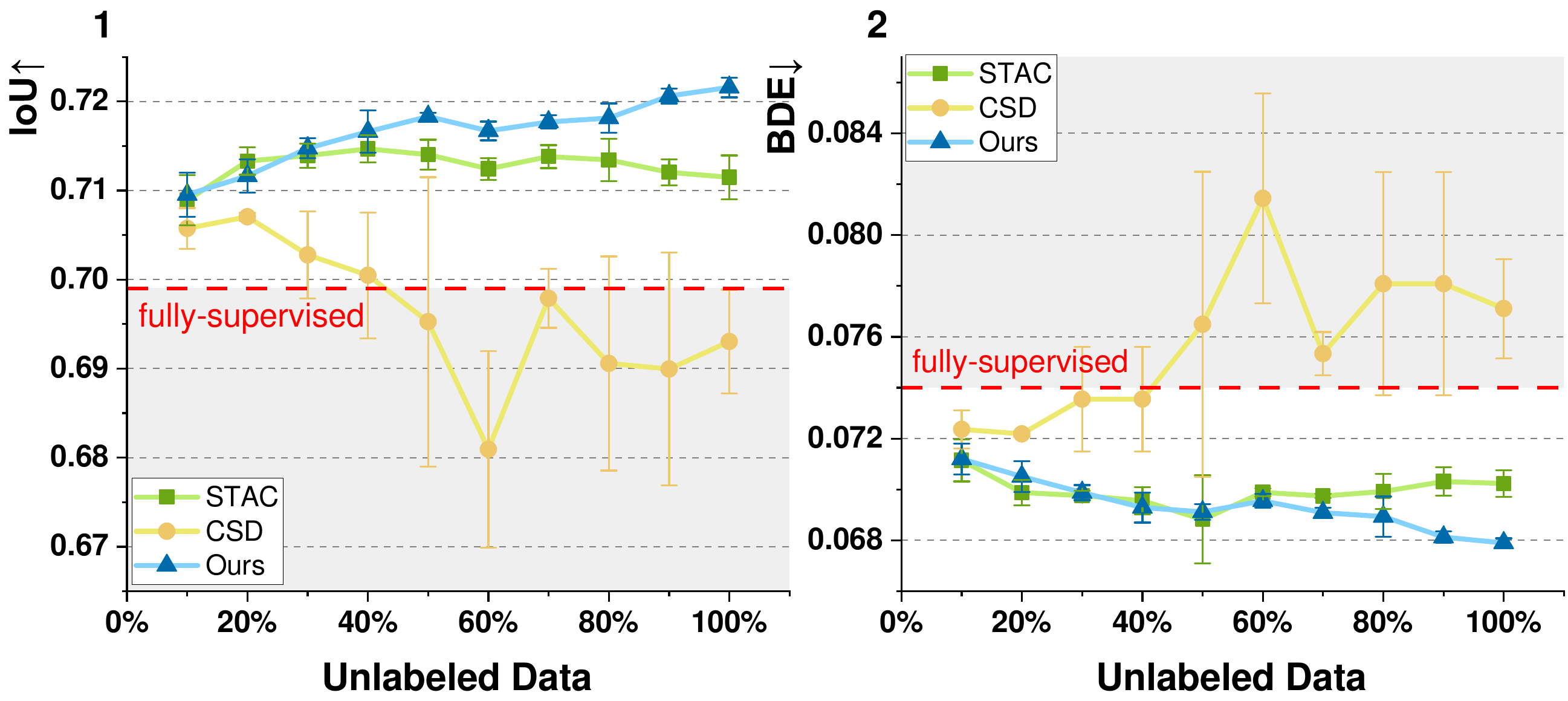}}
  \subcaptionbox{Influence of the unlabeled data quantity on FLMS dataset.}{\includegraphics[width=0.49\textwidth]{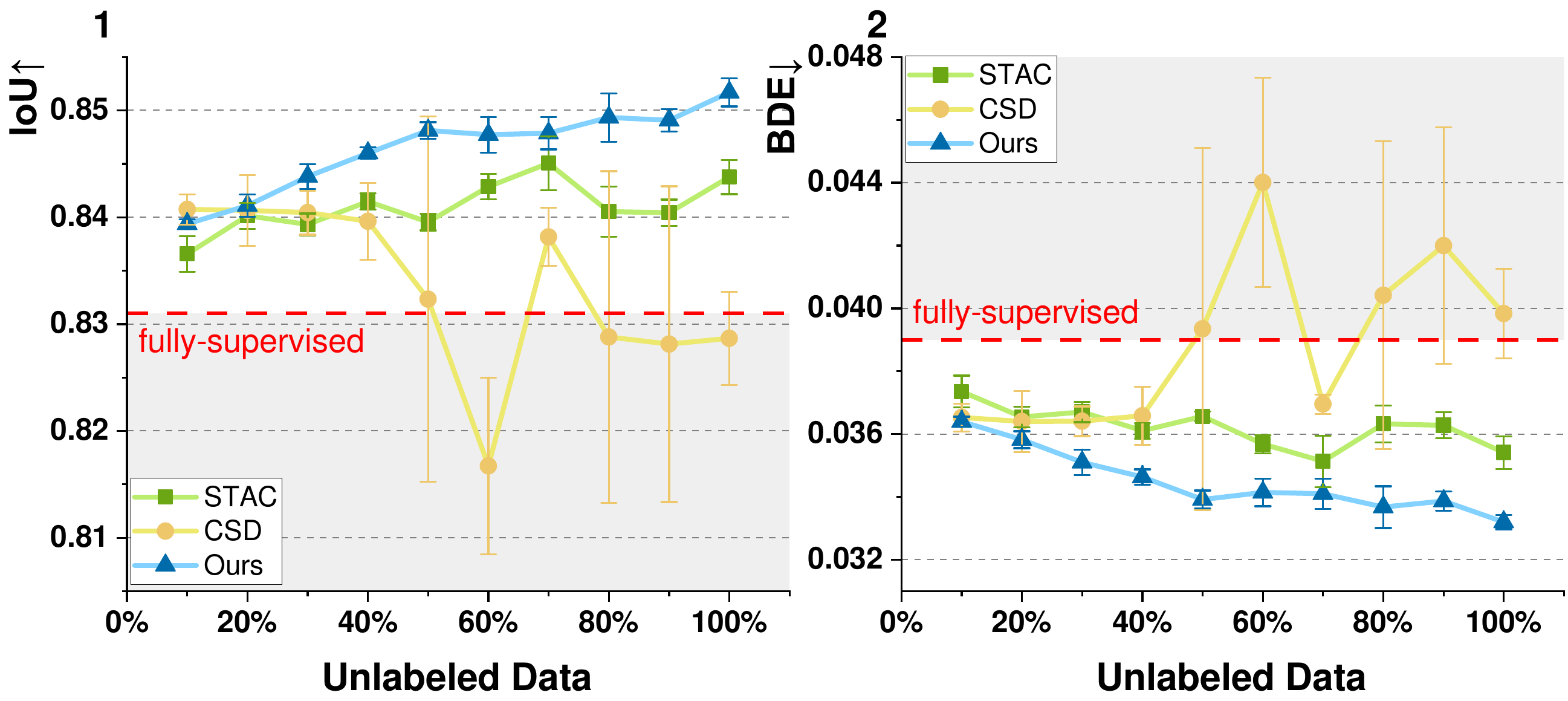}}
  \vspace{-4pt}
  \caption{\textbf{Influence of the unlabeled data quantity.} For each sample point, we repeat the experiment three times and report the mean and standard deviation. The performance of the points in the gray area is worse than the supervised baseline whose performance is indicated by the red dashed line.}
  \label{fig:unlabel}
  \vspace{-10pt}
\end{figure}

\textbf{Influence of the unlabeled data quantity.} We evaluate our framework and compare with typical self-training (STAC) and consistency-regularization (CSD) baselines while varying the amount of unlabeled data. The results are illustrated in Fig.~\ref{fig:unlabel}. One can see that our method outperforms other baselines on nearly all the scales of unlabeled data. Besides, the performance of other baselines cannot benefit from the growth quantity of unlabeled data. In our opinion, for the STAC, this limitation may come from the confirmation bias. When STAC only generates the pseudo labels once and never updates them, more unlabeled data might increase the number of poor pseudo labels. This might make the STAC suffer from more severe confirmation bias. As for the CSD, we think the consistency constraint is too loose. The model may generate poor but consistent recomposed results for the flipped image pairs. Compared with these baselines, our method can perform better when more unlabeled images are available, which demonstrates the potential of our approach for training on larger data scales. Analysis about other influencing factors can be found in supplementary materials.

\begin{table*}[!t]
    \scriptsize
    \centering
    \caption{Ablation experiments. MT, MPV, and PS represent the mean teacher, the MPV-Net, and the policy selecting. Relative performance improvement and decline compared with the supervised baseline (row \romannumeral1) are denoted by the color of \textcolor{blue}{blue} and \textcolor{red}{red}, respectively. ``S'' and ``OS'' represent the supervised learning (training with only labeled data) and omni-supervised (learning training with both labeled and unlabeled data), respectively.}
    \vspace{-6pt}
    \begin{tabular}{c|c|ccc|cccc}
    \toprule
    \multirow{2}*{Index} &\multirow{2}*{Setting} &\multirow{2}*{MT} &\multirow{2}*{MPV} &\multirow{2}*{PS} &\multicolumn{2}{c}{FCDB} &\multicolumn{2}{c}{FLMS}\\
    & & & & & IoU$\uparrow$ &BDE$\downarrow$ &IoU$\uparrow$ &BDE$\downarrow$\\
    \hline
    $\romannumeral1$ &\multirow{3}*{S} &- & & &0.699 &0.074 & 0.831 &0.039\\
    $\romannumeral2$  & &- &\checkmark & &0.702~\textcolor{blue}{(+0.429\%)} &0.074~\textcolor{blue}{(-0.000\%)} & 0.836~\textcolor{blue}{(+0.602\%)} &0.037~\textcolor{blue}{(-5.128\%)}\\
    $\romannumeral3$  & &- &\checkmark &\checkmark &0.704~\textcolor{blue}{(+0.715\%)} &0.073~\textcolor{blue}{(-1.351\%)} & 0.839~\textcolor{blue}{(+0.963\%)} &0.036~\textcolor{blue}{(-7.692\%)}\\
    \hline
     $\romannumeral4$ &\multirow{3}*{OS} &\checkmark & & &0.688~\textcolor{red}{(-1.574\%)} &0.076~\textcolor{red}{(+2.703\%)} &0.834~\textcolor{blue}{(+0.361\%)} &0.037~\textcolor{blue}{(-5.128\%)}\\
    $\romannumeral5$ & &\checkmark &\checkmark & &0.718~\textcolor{blue}{(+2.718\%)} &0.069~\textcolor{blue}{(-6.757\%)} &0.847~\textcolor{blue}{(+1.925\%)} &0.034~\textcolor{blue}{(-12.821\%)}\\
    $\romannumeral6$ & &\checkmark &\checkmark &\checkmark &0.724~\textcolor{blue}{(+3.577\%)} &0.068~\textcolor{blue}{(-8.108\%)} &0.854~\textcolor{blue}{(+2.768\%)} &0.032~\textcolor{blue}{(-17.949\%)}\\
    \bottomrule
    \end{tabular}
    \label{tab:ablation}
    \vspace{-6pt}
\end{table*}

\begin{figure}[!t]
  \centering
  \includegraphics[width=\columnwidth]{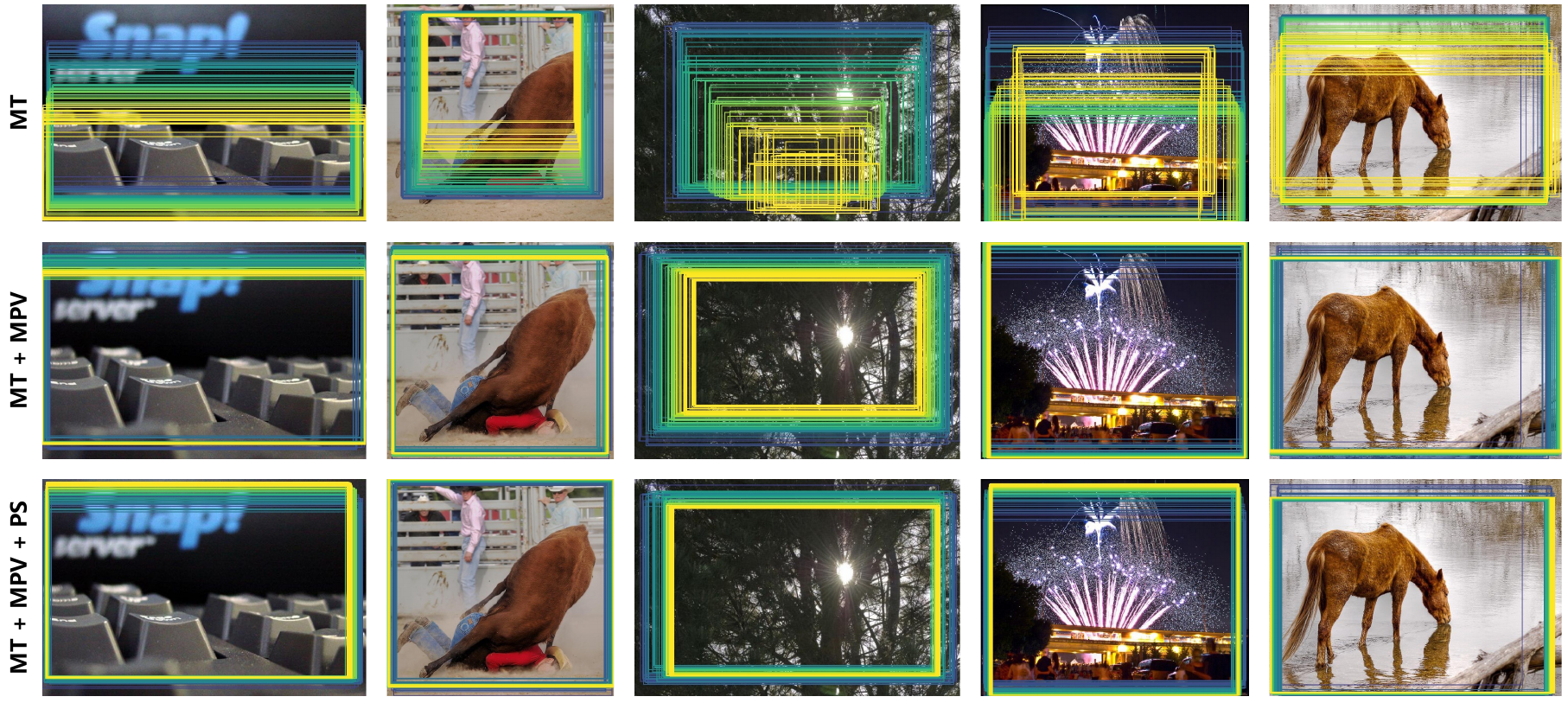}
  \vspace{-16pt}
  \caption{\textbf{The evolution trajectories of pseudo labels for unlabeled data on different ablation settings.} Boxes in different colors are the pseudo labels from different epochs. The \textcolor{light}{light boxes} are the latest pseudo labels; the \textcolor{dark}{dark boxes} are the early generated ones. Best viewed in color.}
  \label{fig:ablation}
  \vspace{-14pt}
\end{figure}

\textbf{Ablation study about mean teacher, MPV-Net, and policy selecting.} 
The quantitative results are shown in Table~\ref{tab:ablation}. 
The setting of supervised learning means training only with labeled data, and the setting of omni-supervised learning means training with both labeled and unlabeled data. Mean teacher~\cite{tarvainen2017mean} is a widely used pseudo-labeling method, which is the baseline of our method.
In the omni-supervised setting, without both the MPV-Net and the policy selecting, our framework degrades to the vanilla mean teacher (row $\romannumeral4$). With the MPV-Net but without the policy selecting (row $\romannumeral5$), we simply fuse the rectified results from different heads of MPV-Net by averaging. We can observe that the performance of the vanilla mean teacher framework drops even lower than the supervised baseline. Adding the MPV-Net can dramatically improve the performance. With the policy selecting (row $\romannumeral6$), the performance is further improved. From the visualization of pseudo labels from different epochs in Fig.~\ref{fig:ablation}, we find that the performance drop of the vanilla mean teacher is caused by the deterioration of the quality of pseudo labels. And the MPV-Net can obviously improve the quality of pseudo labels. However, naive ensembling the policies in MPV-Net still causes some limitations, \eg, too much redundancy and the head-cutting-out mistakes. Policy selecting strategy can alleviate these limitations. These results demonstrate the importance of our contributions. Additional pseudo label visualization can be found in supplementary materials.


\textbf{Self knowledge distillation.} For inference, we can solely use the composer or use both of the composer and MPV-Net. As illustrated in Table~\ref{tab:com2}, training with our MPV-Net and policy selecting, the inference performance with or without MPV-Net are similar; Training without our contributions, the composer cannot perform well. This does not mean MPV-Net and policy selecting are useless. This means that the knowledge of MPV-Net and policy selecting can be distilled to the composer through the omni-supervised learning. Only using the composer which is trained with MPV-Net and policy selecting for inference brings no extra inference cost, which can be seen as an advantage of our method. These results also demonstrate that performance gain come from the self knowledge distillation rather than extra inference cost.

\begin{table}[!t] 
    \small
    \centering
    \caption{Comparison of different inference settings. MPV and PS represent the MPV-Net and the policy selecting, respectively.}
    \vspace{-6pt}
    \resizebox{0.75\linewidth}{!}{
    \begin{tabular}{ccccccccc}
    \toprule
    \multirow{2}*{Inference} &Trained with &Extra  &\multicolumn{2}{c}{FCDB} &\multicolumn{2}{c}{FLMS} &\multirow{2}*{FPS}\\
    &MPV+PS &Annotation & IoU$\uparrow$ &BDE$\downarrow$ &IoU$\uparrow$ &BDE$\downarrow$\\
    \hline
    CACNet~\cite{hong2021composing} &- &$\checkmark$ &0.716 &0.069 &0.853 &0.033 &150\\
    \hline
     composer &$\times$ &$\times$  &0.686 &0.076 &0.832 &0.038 &200\\
     composer &$\checkmark$ &$\times$ &0.723 &0.068 &0.853 &0.033 &200\\
     composer+MPV-Net+PS &$\checkmark$ &$\times$  &0.724  &0.068 &0.854 &0.032 &88\\
    \bottomrule
    \end{tabular}}
    \label{tab:com2}
    \vspace{-10pt}
\end{table}

\section{Conclusion}

For the first time, we introduce pseudo-labeling into image cropping task, which makes it possible to use both the labeled and unlabeled data. Learning aesthetic composition patterns is the core of this task. Human beings acquire knowledge of aesthetics by constantly being exposed to masterpieces, which can be considered as labeled data, as well as a large amount of works from ordinary people, which can be treated as unlabeled data. From this perspective, aesthetic cognition of human-beings seems to derive from an omni-supervised learning paradigm~\cite{zhu2007humans}. However, learning-based image cropping methods are dominated by supervised ones.
This research gap comes from that the unquantifiable aesthetic quality makes it challenging to find better pseudo labels. So, the confirmation bias of the self-training framework is difficult to be addressed. We fill this gap by hunting for better pseudo labels with the MPV-Net and policy selecting. Experimental results illustrate a large performance gain over the supervised baseline, and a obvious performance improvement compared with other SOTA image cropping solutions that even uses extra labeled data. We hope our work can shift the attention of the image cropping community from supervised learning to omni-supervised paradigm.

%
%
\bibliographystyle{splncs04}
\bibliography{main}
\end{document}